%% file: main.tex
\documentclass[10pt,twocolumn,letterpaper]{article}

\usepackage{cvpr}              
\usepackage{svg}
\usepackage{float}


\definecolor{cvprblue}{rgb}{0.21,0.49,0.74}
\usepackage[pagebackref,breaklinks,colorlinks,allcolors=cvprblue]{hyperref}

\def\paperID{13} 
\def\confName{CVPR}
\def\confYear{2025}

\title{Cycle Training with Semi-Supervised Domain Adaptation: Bridging Accuracy and Efficiency for Real-Time Mobile Scene Detection}

\author{
\thanks{%
    All authors contributed equally to this paper.\\
    This research is fully supported by AI VIETNAM \cite{aivietnamVit}.%
  }
  Huu-Phong Phan-Nguyen $^{1*}$ \quad Anh Dao$^{2*}$ \quad  Tien-Huy Nguyen$^{1*}$ \\ \quad Tuan Quang$^{3}$  \quad Huu-Loc Tran$^{1}$ \quad Tinh-Anh Nguyen-Nhu$^{4}$ \\ \quad Huy-Thach Pham $^{5}$\quad Quan Nguyen $^{6}$\quad Hoang M. Le $^{7}$\quad Quang-Vinh Dinh $^{8}$\\ \\
$^{1}$ Ho Chi Minh University of Technology, VNU-HCM, Vietnam \\
$^{2}$ Michigan State University, USA \\
$^{3}$ LPL Financial, USA \\
$^{4}$ Ho Chi Minh University of Technology, VNU-HCM, Vietnam \\
$^{5}$  Georgia State University, USA \\
$^{6}$ Posts and Telecommunications Institute of Technology, Hanoi, Vietnam \\
$^{7}$ York University, Canada\\
$^{8}$ AI VIETNAM Lab \\
{\tt\small 22520567@gm.uit.edu.vn}}

\begin{document}
\maketitle
\input{sec/0_abstract}    
\input{sec/1_intro}

\input{sec/2_related}
\input{sec/3_method}

\input{sec/4_experiment}
\input{sec/5_conclusion}

{
    \small
    \bibliographystyle{ieeenat_fullname}
    \bibliography{main}
}


\end{document}

%% file: sec/0_abstract.tex
\begin{abstract}
Nowadays, smartphones are ubiquitous, and almost everyone owns one. At the same time, the rapid development of AI has spurred extensive research on applying deep learning techniques to image classification. However, due to the limited resources available on mobile devices, significant challenges remain in balancing accuracy with computational efficiency. In this paper, we propose a novel training framework called Cycle Training, which adopts a three-stage training process that alternates between exploration and stabilization phases to optimize model performance. Additionally, we incorporate Semi-Supervised Domain Adaptation (SSDA) to leverage the power of large models and unlabeled data, thereby effectively expanding the training dataset. Comprehensive experiments on the CamSSD dataset for mobile scene detection demonstrate that our framework not only significantly improves classification accuracy but also ensures real-time inference efficiency. Specifically, our method achieves a \textbf{94.00\%} in Top-1 accuracy and a \textbf{99.17\%} in Top-3 accuracy and runs inference in just \textbf{1.61ms} using CPU, demonstrating its suitability for real-world mobile deployment. 
\end{abstract}

%% file: sec/1_intro.tex
\section{Introduction}
\label{sec:intro}

Deep learning show exceptional abilities across diverse tasks such as visual question answering, object detection, image retrieval, domain adaption, and recognition \cite{10.1145/3628797.3629011, 10661057, nguyen2024improvinggeneralizationvisualreasoning, nguyensemi, nguyen2024emotic, nguyen2025enhancing, ngo2024dual} but has traditionally been associated with high-performance computing environments such as data centers and cloud servers. However, the emerging demand for advanced deep learning models running on resource-constrained devices—such as smartphones, embedded systems, and microcontrollers—has led to significant research interest in what is often referred to as edge AI or tiny machine learning (TinyML) \cite{tinyml1, tinyml2, tinyml3}. These small devices are defined by their constrained computational resources, limited memory, and strict power consumption requirements. For example, while modern smartphones integrate increasingly powerful processors due to Moore’s Law, they still face inherent limitations compared to server-grade hardware. Embedded systems, such as those in IoT devices, and microcontrollers, like those in Arduino or STM32 boards, operate under even tighter constraints, often running on real-time operating systems (RTOS) or bare-metal environments with minimal available memory.

The challenge of deploying deep learning models on edge devices stems from the computational demands of modern neural networks. Deep models require substantial processing power for inference, large memory footprints to store millions of parameters, and sustained energy availability, which is a crucial factor for battery-operated devices. For instance, state-of-the-art convolutional neural networks (CNNs) for image recognition can exceed 100 million parameters, making them impractical for deployment on devices with only a few megabytes of RAM and storage. Additionally, power efficiency is critical, as prolonged deep learning inference can rapidly drain the battery of mobile devices.

To overcome these challenges, various techniques have been researched and developed to optimize the size of the model for deployment on resource-constrained devices while maintaining competitive performance comparable to larger models. In terms of model compression, the pruning method \cite{han2015learning, han2015deep, he2017channel} removes unnecessary or low-impact connections in a neural network, significantly reducing model size while preserving accuracy. The key idea is that many connections in a trained model contribute little to the performance and can be eliminated without recording any significant loss. The quantization technique like post-training quantization and quantization-aware training \cite{krishnamoorthi1806quantizing, han2015deep} has also been a common choice in model compression, as it reduces the precision of model weights and activations (e.g., from 32-bit floating point to 8-bit integers), leading to faster computation and lower memory usage. However, to improve overall performance with limited size, knowledge distillation \cite{hinton2015distilling} has been used widely, as it transfers knowledge from a large and high performing "teacher" model to a smaller "student" model. The student learns from the softened probability distributions produced by the teacher, allowing it to maintain high accuracy while being significantly more efficient. Advancements in model architecture have been instrumental in enhancing the efficiency of deep learning models for mobile and edge devices. The incorporation of depthwise separable convolutions in the MobileNet series \cite{dong2020mobilenetv2, koonce2021mobilenetv3, qin2024mobilenetv4} and the introduction of compound scaling in EfficientNet \cite{ab2021efficientnet, koonce2021efficientnet},  have significantly improved model performance while reducing computational complexity. These innovations enable models to achieve a superior balance between accuracy and efficiency, making them well-suited for real-time applications on resource-constrained devices. Despite these advancements, achieving high accuracy, fast inference speed, and minimal memory consumption simultaneously remains a challenge. Many optimized models still suffer from accuracy degradation due to aggressive compression or struggle with domain adaptation when trained in one environment but deployed in another.
To address these issues, we propose a novel framework that enhances model efficiency and generalization on resource-constrained devices while ensuring high performance. Our main contributions are:
\begin{itemize}
    \item \textbf{Semi-supervised Domain Adaptation (SSDA)} that takes advantage of high-confidence pseudo-labeled data to mitigate the domain gap between training and testing data, effectively expanding the available labeled dataset.

    \item \textbf{Cycle Training (CT)} strategy that iteratively refines the student model by alternating between full fine-tuning (exploration) and selective parameter updates (stabilization), thereby ensuring robust knowledge transfer from a high-capacity teacher network.

    \item To validate the effectiveness of our proposed method, we conduct a comprehensive suite of experiments on a challenging mobile scene detection dataset, demonstrating significant improvements in both accuracy and inference speed on mobile devices.
\end{itemize}

%% file: sec/2_related.tex
\section{Related Work}
\label{sec:related}

\textbf{Knowledge Distillation}. Knowledge distillation (KD) \cite{gou2021knowledge, know2, know3} has emerged as a powerful technique for compressing large neural networks into smaller, more efficient models suitable for resource-constrained environments. Initially proposed by Hinton et al.\cite{hinton2015distilling}, KD transfers knowledge from a large and trained teacher model to a lightweight student model, enabling deployment on devices with limited computational power and memory. The method leverages soft targets from teacher networks and has been widely adopted for compressing neural networks. In speech recognition, a cornerstone of mobile device functionality (e.g., voice assistants), has benefited significantly from KD. Li et al. \cite{yang2019essence} propose distilling knowledge from an ensemble of acoustic models into a single compact model, reducing computational overhead for real-time speech processing on mobile devices. Similarly, Chebotar and Waters \cite{chebotar2016distilling} demonstrate how KD can improve accuracy while maintaining a lightweight architecture that is suitable for mobile development. Beyond speech, KD has been explored for other mobile-centric tasks. Alkhulaifi et al. \cite{alkhulaifi2021knowledge} discussed its use in human activity recognition on smartphones, leveraging sensor data to train compact models for on-device inference. This work cites previous efforts \cite{plotz2018deep} that distill cloud-trained models for mobile development, showcasing the versatility of KD in mobile applications. Subsequent surveys \cite{gou2021knowledge, moslemi2024survey} provide comprehensive overviews of KD techniques, highlighting their applicability to mobile devices and embedded systems. These works highlight the balance and size of the model, which are crucial for real-time mobile applications.

\textbf{Model architecture}. Recent advances in deep learning have enabled real-time scene detection on mobile devices by leveraging lightweight neural network architectures. Traditional deep learning models \cite{alexnet, densetnet, inception}, such as ResNet \cite{resnet} or VGG \cite{vggnet}, are computationally expensive and unsuitable for mobile deployment due to limited processing power and battery constraints. To address this, researchers have explored small-scale models \cite{howard2017mobilenets, han2020ghostnet, dong2020mobilenetv2, koonce2021mobilenetv3, qin2024mobilenetv4} optimized for efficiency while maintaining high accuracy. An widely adopted approach is the use of MobileNetV2 \cite{dong2020mobilenetv2}, MobileNetV3 \cite{koonce2021mobilenetv3}, and the recently proposed MobileNetV4 \cite{qin2024mobilenetv4}, which utilize advanced techniques to reduce computation while maintaining feature extraction capabilities. MobileNetV2 \cite{dong2020mobilenetv2} introduced depthwise separable convolutions, significantly reducing FLOPs. MobileNetV3 \cite{koonce2021mobilenetv3} further optimized the architecture with squeeze-and-excitation (SE) modules and neural architecture search (NAS), achieving a balance between latency and accuracy. MobileNetV4 \cite{qin2024mobilenetv4}, an evolution of its predecessors, integrates efficient self-attention mechanisms and dynamic convolutions to improve accuracy while maintaining low computational cost, making it highly suitable for real-time applications. Other lightweight architectures such as ShuffleNet \cite{hluchyj1991shuffle} and EfficientNet-Lite \cite{ab2021efficientnet} employ channel shuffling and compound scaling strategies, respectively, to further enhance efficiency. In addition, quantization and model pruning techniques have been widely used to compress models for mobile deployment. TensorFlow Lite and PyTorch Mobile offer post-training quantization tools that allow size reduction without significant accuracy loss. Deep compression methods, such as those proposed by \cite{han2016deep}, further optimize models through pruning and weight sharing.

\textbf{Semi-Supervised Learning}. Semi-supervised learning (SSL) leverages both labeled and unlabeled data to enhance the model performance, a critical approach in computer vision, where labeled data are often scarce. Recent advances have refined SSL for image classification and related tasks. FixMatch \cite{sohn2020fixmatch}, which combines consistency regularization and pseudo-labeling with a fixed confidence threshold. FlexMatch \cite{zhang2021flexmatch} introduces Curriculum Pseudo Labeling, where it adatively adjusts thresholds based on class-specific learning status, enhancing performance on imbalanced datasets. FreeMatch \cite{wang2022freematch} further advances this with self-adaptive thresholding and class fairness regularization, reducing error rates for rare labeled data scenarios. FlatMatch \cite{huang2023flatmatch} focuses on generalization by minimizing cross-sharpness, ensuring consistent learning between labeled and unlabeled data, while RegMixMatch \cite{han2024regmixmatch} optimizes the Mixup  \cite{zhang2017mixup} augmentation technique with semi-supervised and class-aware strategies, improving robustness through data augmentation. These methods have been used widely in competitions in which the validation set is treated as an unlabeled dataset, further enhancing the performance of the model. 

%% file: sec/3_method.tex
\section{Method}
\label{sec:method}
\subsection{Overview}

\begin{figure*}
    \centering
    \includegraphics[width=0.85\textwidth]{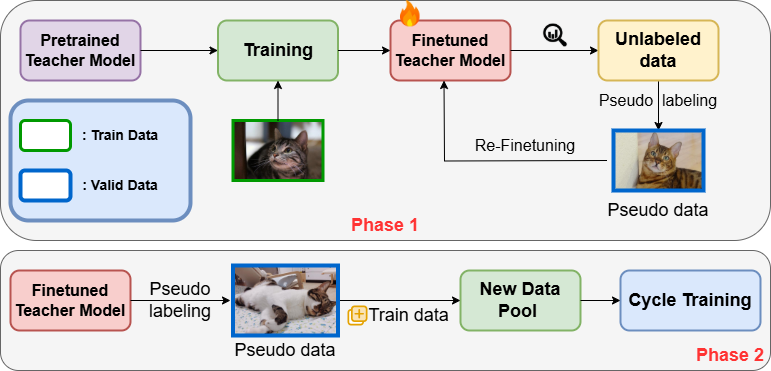}
    \caption{Overview of our architecture. In the first phase, the pretrained teacher model is fine-tuned using labeled training data. It then applies Semi-Supervised Domain Adaptation illustrated in \textbf{Section}~\ref{subsec:pseudo} to predicts unlabeled data to generate pseudo labels, followed by re-finetuning. In the second phase, the fine-tuned teacher model continues generating pseudo-labeled data, which is combined with the original training data to form a new dataset. This dataset is then used for Cycle Training (described in \textbf{Section}~\ref{subsec:stage}) the student model.}
    \label{fig:overview}
\end{figure*}


\label{subsec:overview}

In real-time mobile scene detection, model size is an important factor as it involves a trade-off between performance and the computational and memory constraints of mobile devices. Our approach as illustrated in \textbf{Figure \ref{fig:overview}}  is driven by the need to deploy efficient an accurate models on resource-limited hardware. Our key idea is to capitalize on the strengths of large-scale models during training while ensuring that the final model remains lightweight enough for real-time applications.

As illustrated in \textbf{Figure}~\ref{fig:overview}, the method begins by leveraging a high-capacity network, which is first trained on a fully labeled dataset (train dataset). In our approach, we utilize ResNet-101x3 as the teacher model for pseudo labeling. This large model is adept at capturing intricate feature representations and provides reliable predictions. These predictions are then used to generate pseudo labels for additional, unlabeled data (val data). These pseudo labels are leveraged to fine-tune the ResNet further, enhancing its overall performance. With the refined ResNet model, we generate a new set of pseudo labels that are more accurate. By combining the original labeled data with high-confidence pseudo labels from the improved ResNet, we train a lightweight MobileNetv2 model using Cycle Training. This strategy enables the compact MobileNetv2 to inherit robust features from the larger ResNet while ensuring the efficiency needed for real-time mobile scene detection. 

\textbf{ResNet-101x3 \cite{kolesnikov2020big}- Teacher model.} ResNet-101, as employed in the Big Transfer (BiT) framework, benefits from extensive pre-training on massive, diverse datasets, endowing it with highly transferable feature representations. In our modified architecture, we replace the original classification head with a single fully connected layer that directly maps the features extracted from the last convolutional layer to the 30 target classes. This design enables effective fine-tuning on our specific dataset, ensuring robust performance in real-time mobile scene detection.

\textbf{MobileNet-V2 \cite{sandler2018mobilenetv2} - Student model.} MobileNet-V2 is designed for efficiency on mobile and embedded devices, offering a favorable balance between computational cost and accuracy. It utilizes depthwise separable convolutions to significantly reduce the number of parameters while maintaining effective feature extraction. For MobileNet-V2, we substitute the original head with a two-layer fully connected structure. The first fully connected layer comprises 1280 units and employs the ReLU activation function to effectively capture high-level feature abstractions. This is followed by a second fully connected layer that projects the 1280-dimensional feature vector onto the 30 output classes. A Softmax activation is then applied to the final layer to obtain the class probabilities, ensuring the model is both efficient and accurate for real-time mobile scene detection.

\subsection{Semi-Supervised Domain Adaptation }
\label{subsec:pseudo}
Owing to the limited availability of data with precise labels, training a model solely on the samples found in the training set can lead to overfitting due to the inherent domain gap between the training and testing sets. To overcome this challenge and address the potential domain bias separating the two sets —which can give rise to unobserved scenarios during testing — we propose a Semi-Supervised Domain-Adaptive (SSDA) training strategy. SSDA, a semi-supervised learning technique that enhances model performance by incorporating high-confidence pseudo labels into the training process. This approach effectively expands the labeled dataset by utilizing the model’s own predictions as additional training samples, allowing it to learn from both labeled and unlabeled data. 

In our framework, once the initial model is trained on the fully labeled dataset, it is employed to predict labels for unlabeled samples. By enforcing a strict confidence threshold, only predictions with sufficiently high certainty are selected as pseudo labels. This selection mechanism ensures that only reliable predictions augment the training set, minimizing the risk of incorporating noisy or incorrect labels. Consequently, the expanded data set, which now includes both original labels and pseudo-labeled models, provides a richer and more diverse set of training examples. 

For each unlabeled sample \( x \), the model produces a probability distribution \( f(x) \) over the target classes. We define the confidence score for \( x \) as:

\begin{equation}
s(x) = \max_{c} f(x)_c
\end{equation}
where \( f(x)_c \) denotes the predicted probability for class \( c \). A pseudo label \( y_p \) is then assigned to \( x \) if its confidence score meets or exceeds a predefined threshold \( \tau \) (e.g., 0.8):
\begin{equation}
    y_p =
\begin{cases}
\arg\max_{c} f(x)_c, & \text{if } s(x) \geq \tau, \\
\text{discarded}, & \text{otherwise}.
\end{cases}
\end{equation}

Only samples with \( s(x) \geq \tau \) are retained and added to the training set. The model is subsequently fine-tuned on this expanded dataset, leading to improved generalization and enhanced performance.

\subsection{Data Augmentation}
\label{ssubsec:augment}
To improve the robustness of our MobileNetV2 \cite{sandler2018mobilenetv2} model, we employed a combination of weak and strong data augmentation techniques. Initially, all input images are subjected to random cropping to ensure spatial diversity. For each image \( x \), a random probability \( p \) is drawn from a uniform distribution:
\begin{equation}
    p \sim \mathcal{U}(0,1).
\end{equation}
Based on the value of \( p \), the augmentation applied to \( x \) is defined as:

\begin{equation}
        \text{Augment}(x) =
\begin{cases} 
\text{Flip}(x), & \text{if } p < 0.3, \\[1mm]
\text{RanAug}(x, n, m), & \text{if } p > 0.7, \\[1mm]
x, & \text{otherwise}.
\end{cases}
\end{equation}

Here, \(\text{Flip}(x)\) denotes a horizontal flip operation, while \(\text{RanAug}(x, n, m)\) applies a sequence of \( n \) randomly selected transformations with an intensity level \( m \).

For strong augmentation, we adopt the \emph{RanAug} strategy, which selects augmentation operations from the set
\[
\mathcal{A} = \{\text{AutoContrast},\, \text{Brightness},\, \text{Color},\, \text{Contrast},\, \text{Rotate}\}.
\]
In our experiments, we set the parameters to \( n = 2 \) (the number of augmentation operations per image) and \( m = 5 \) (where \( m \in [0,10] \) controls the magnitude of each transformation).

Furthermore, a cut-off mechanism is employed during the application of RandomAugment to ensure that the intensity of the augmentations remains within a reasonable range, preventing excessive perturbations.

This augmentation pipeline combines weak and strong techniques, enhancing the model's generalization ability by introducing diverse variations while preserving the essential features of the images.

\begin{figure*}
    \centering
    \includegraphics[scale=0.85]{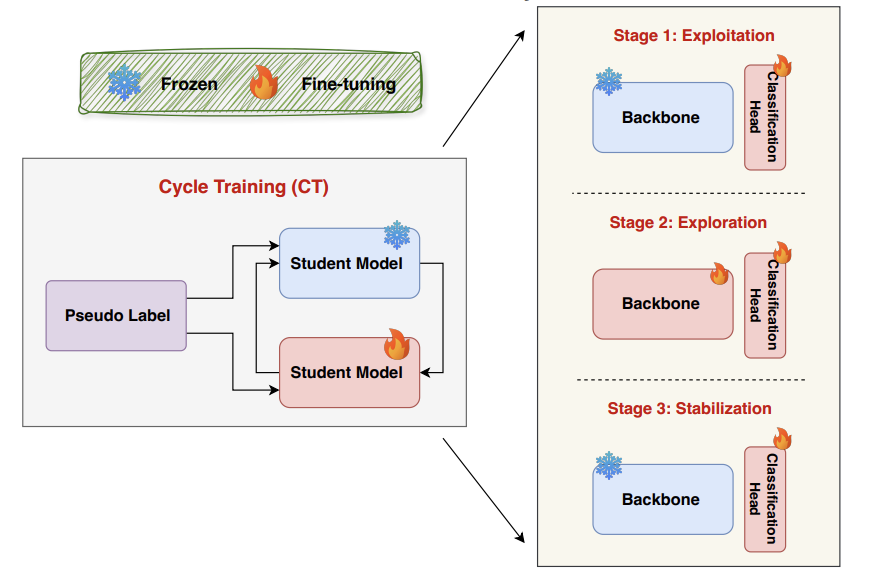}
    \caption{Illustration of the proposed Cycle Training (CT) framework. The process consists of three progressive stages: (1) Exploitation, where only the classification head is fine-tuned while keeping the backbone frozen; (2) Exploration, where the entire student model is fine-tuned using pseudo-labeled data; and (3) Stabilization, where the backbone is frozen again while refining the classification head. This cyclic training strategy helps balance knowledge preservation, exploration, and stabilization, ensuring effective knowledge transfer from the teacher model to the student model.}
    \label{fig:CT}
\end{figure*}

\subsection{Cycle Training (CT)}
The primary objective of this section is to extract and transfer all the knowledge embedded in the teacher model into the student model while preserving its representational power and generalizability. Instead of merely fine-tuning the model on the target dataset, we propose a progressing \textbf{Cycle Training (CT)}  as described in \textbf{Figure} \ref{fig:CT} framework, a structured multi-stage optimization process that iteratively refines the student model through staged fine-tuning and pseudo-labeling. Our method assumes that the distilled knowledge is progressively integrated into the student networks without catastrophic forgetting or overfitting to the pseudo-labeled data.

The proposed framework begins with training a high-capacity ResNet-101x3\cite{kolesnikov2020big} model, which serves as a feature-rich teacher network. Once trained, this model is leveraged to generate pseudo-labels on the validation set. These pseudo-labels are then combined with the original dataset to create an augmented dataset used for knowledge transfer. However, directly fine-tuning this student model often leads to suboptimal convergence and loss of crucial representations. To mitigate this, we introduce a three-stage progressive training strategy that balances knowledge preservation, exploration, and stabilization. 

In\textbf{ Stage 1 (Exploitation Phase)}, the student's backbone remains frozen, and only the classification head is fine-tuned for 10 epochs. This initial phase allows the model to absorb task-specific features without disrupting the pre-trained feature representations. By restricting only the gradient updates to the classification head, the model can retain the hierarchical features it learns from its pretraining phase while quickly adapting to the new dataset. Following this, in \textbf{Stage 2 (Exploration Phase)}, the entire student model—including the backbone—is fully fine-tuned on the augmented dataset. This phase allows the student model to explore new feature representations while leveraging the information embedded in the pseudo-labels. Unlike direct end-to-end fine-tuning, which often leads to overfitting due to noisy pseudo-labels, our final stage ensures stable adaptation to the new data distribution. Finally, in \textbf{Stage 3 (Stabilization Phase)}, the backbone is frozen again while the classification head undergoes additional fine-tuning. This mechanism serves as a regularization mechanism, preventing the model from overfitting while reinforcing the learned feature representations.

The main key motivation of this cyclic approach is to maximize the transfer of knowledge while mitigating overfitting due to noisy labels in pseudo-labeling. Our approach actively regulates the learning process through alternating stability and exploration phases. The whole training paradigm closely resembles the principles of simulated annealing, where an initial structured phase prevents premature convergence, followed by exploration to reach an optimal solution, and the final refinement phase to stabilize the learning trajectory, prevent it from escape the ideal solution. With this approach, the student model gradually inherits the knowledge of the teacher network, while preserving its computational efficiency.

\label{subsec:stage}

%% file: sec/4_experiment.tex
\section{Experiment}
\label{sec:experiment}

In this section, we present a comprehensive evaluation of our proposed method on the Mobile AI Workshop dataset for real-time camera scene detection. Additionally, we provide a detailed description of the dataset, evaluation metric, implementation details, quantitative, quanlitative results and extensive ablation experiments.


\textbf{Dataset:} In the MAI workshop challenge, we use the Camera Scene Detection Dataset (CamSDD) \cite{pouget2021fast}. It comprises 30 classes with over 11K images. The class distribution is relatively balanced, with each category containing approximately 330 images on average. Images are resized to a resolution of 576x284 pixels, striking a balance between computational efficiency and visual detail suitable for mobile applications.


\textbf{Evaluation Metrics: } The evaluation of the model performance in this challenge is based on two key aspects: (1) \textbf{prediction accuracy}, measured using top-1 and top-3 classification accuracy on the test set, and (2) \textbf{runtime efficiency}, quantified based on inference latency on the actual target mobile platform. The final ranking of submissions is determined by a composite score that balances accuracy and runtime:

\begin{equation}
    Score(Top1, Top3,runtime)= \frac{2(Top1+Top3)}{C \cdot runtime}
\end{equation}

\begin{table*}[!ht]
    \centering
    \renewcommand{\arraystretch}{1.2} 
    \setlength{\tabcolsep}{8pt} 
    \begin{tabular}{l c c c c}
        \toprule
        Team & Model Size (MB) & Top-1 (\%) & Top-3 (\%) & Final Runtime (ms) \\
        \midrule
        EVAI & 0.83 & 93.00 & 98.00 & 3.35 \\
        \textit{MobileNet-V2} & 18.6 & 94.17 & 98.67 & 16.38 \\
        ALONG & 12.7 & 94.67 & 99.50 & 64.45 \\
        Team Horizon & 2.27 & 92.33 & 98.67 & 7.7 \\
        Airia-Det & 1.36 & 93.00 & 99.00 & 17.51 \\
        DataArt Perceptrons & 2.73 & 91.50 & 97.67 & 54.13 \\
        PyImageSearch & 2.02 & 89.67 & 97.83 & 45.88 \\
        neptuneai & 0.045 & 83.67 & 94.67 & 4.17 \\
        Sidiki & 0.072 & 78.00 & 93.83 & 1.74 \\
        \textbf{GenAI4E (Ours)} & 14.85 & 94.00 & 99.17 & \textbf{1.61} \\
        \bottomrule
    \end{tabular}
    \caption{Comparison of model performance metrics, including model size, accuracy (Top-1 and Top-3), and final runtime \cite{mobile2}. The proposed GenAI4E (Ours) model achieves a strong balance between accuracy and efficiency, demonstrating the lowest inference time while maintaining competitive accuracy.}
    \label{tab:model_comparison}
    \vspace{-1.5em}
\end{table*}

where $C$ is a normalized constant. While top-ranked models are determined based on this score, models exhibiting exceptional efficiency-accuracy trade-offs are also granted awards.

The evaluation is conducted on image pairs (low- and high-resolution) across 30 camera scene classes. Prediction accuracy is measured using top-1 and top-3 metrics on the test set, while runtime is recorded on a target mobile device (e.g., Samsung Galaxy S10 with NNAPI acceleration). All evaluation scripts and TFLite conversion tools are provided to ensure full reproducibility.

\textbf{Implementation In Details: } We separate training configurations into two different ones for each model like that: 

\begin{itemize}
    \item \textbf{ResNet-101x3 \cite{kolesnikov2020big} - Teacher Network and Pseudo Labeling:}  We first resize the input images to 256x256 and then perform a random crop to obtain 224x224 images. The model is trained using the AdamW optimizer with an initial learning rate of \(1.5 \times 10^{-3}\) and a batch size of 256. Training is conducted in two phases: 10 epochs on the labeled training data, followed by 10 epochs on the extra pseudo-labeled data. 
    \item \textbf{Student Network Training via Cycle Training (CT):}  A lightweight MobileNetv2 model is trained within a CT that integrates refined pseudo labels and a tailored data augmentation module. The augmentation strategy employs both weak and strong transformations. We train the model for a total of 50 epochs: 10 epochs for stage 1, 30 epochs for stage 2, and 10 epochs for stage 3, with a batch size of 32. Input images are resized to 160x160, then randomly cropped to 128x128, and normalized to the range [-1, 1] for faster computation. An exponential decay learning rate scheduler is used with an initial learning rate of \(10^{-3}\), decaying by a factor of 0.1 every 20 epochs.
\end{itemize}

\subsection{Quantitative Results}
To assess the impact of pseudo labeling, data augmentation, and the Cycle Training strategy, we conducted a series of ablation studies.

The \textbf{Table}~\ref{tab:model_comparison} presents a comparative analysis of different models based on key performance metrics, including model size (in MB), Top-1 and Top-3 accuracy (in percentage), and final runtime (in milliseconds). The evaluation highlights the trade-offs between model size, accuracy, and computational efficiency across various teams and approaches. The \textbf{GenAI4E (Ours)} model demonstrates a strong balance between accuracy and efficiency. Despite having a model size of 14.85 MB, it achieves a competitive \textbf{94.00\% Top-1 accuracy and 99.17\% Top-3 accuracy}, with an outstanding runtime of \textbf{only 1.61 ms}, making it the most efficient model in terms of inference speed. This result underscores the effectiveness of the proposed method in maintaining high accuracy while optimizing computational efficiency. MobileNet-V2, a well-known lightweight architecture, maintains a competitive Top-1 accuracy of 94.17\% but has a larger model size (18.6 MB) and higher runtime (16.38 ms), indicating a potential trade-off between accuracy and efficiency. Other models, such as Sidiki and neptuneai, have significantly smaller model sizes (0.072 MB and 0.045 MB, respectively), but their lower accuracy scores (78.00\% and 83.67\% in Top-1) suggest that extreme model compression might lead to performance degradation. Conversely, ALONG achieves a high Top-1 accuracy (94.67\%) but at the cost of a large runtime (64.45 ms), demonstrating the potential trade-off between accuracy and computational overhead.

\textbf{Overall}, the results highlight the importance of balancing model size, accuracy, and runtime for optimal performance in real-world applications, with \textbf{GenAI4E (Ours)} showcasing a superior balance between these factors.

\subsection{Ablation Studies}

\begin{table}[ht]
    \centering
    \renewcommand{\arraystretch}{1.2}
    \setlength{\tabcolsep}{6pt}
    \begin{tabular}{l c c c}
        \toprule
        \textbf{Model} & \textbf{Threshold} & \textbf{Top-1 (\%)} & \textbf{Top-3 (\%)} \\
        \midrule
        ResNet101x3 & -- &  97.00 & - \\
        ResNet101x3 & 0.9 & \textbf{97.50} & - \\
        \midrule
        \multicolumn{3}{l}{\textit{Test on validation set}} \\
        \midrule
        MobileNetv2 & -- & 91.67 & 98.83 \\
        MobileNetv2 & 0.0 & 91.33 & 99.00 \\
        MobileNetv2 & 0.8 & \textbf{92.83} & 98.83 \\
        MobileNetv2 & 0.85 & 92.00 & \textbf{99.17} \\
        MobileNetv2 & 0.9 & 92.17 & 98.83 \\
        \midrule
        \multicolumn{3}{l}{\textit{Test on test set}} \\
        \bottomrule
    \end{tabular}
    \caption{Evaluate the impact of threshold in generating pseudo-label when applying Semi-Supervised Domain Adaptation.}
    \label{tab:pseudo}
\end{table}

The results in \textbf{Table \ref{tab:pseudo}} present the impact of pseudo-labeling on the performance of MobileNetV2 and ResNet-101x3. Comparing the performance of ResNet-101x3 before and after applying pseudo-labeling, we observe a slight decrease in Top-1 accuracy (from 97.00\% to 97.50\%). This suggests that while pseudo-labeling can introduce additional training data, it may also incorporate some noise, leading to minor fluctuations in performance. For MobileNetV2, pseudo-labeling improves performance over the baseline (91.67\% Top-1). Without confidence filtering (\(\tau = 0)\), accuracy drops slightly (91.33\%) due to noisy labels. We select (\(\tau = 0.8)\) as the optimal threshold, as it achieves the highest Top-1 accuracy (92.83\%) while maintaining stable Top-3 accuracy (98.83\%). This balance ensures that the model benefits from high-quality pseudo labels while retaining sufficient training data for effective learning. 

\begin{table}[ht]
    \centering
    \footnotesize
    \setlength{\tabcolsep}{4pt}
    \renewcommand{\arraystretch}{1.2} 
    \begin{tabular}{l c c c c c}
        \toprule
        \textbf{Method} & \textbf{Stage 1} & \textbf{Stage 2} & \textbf{Stage 3} & \textbf{Top-1 (\%)} & \textbf{Top-3 (\%)} \\
        \midrule
        Ours & \checkmark   & -- & -- &  89.83 & 97.67 \\
        Ours &  \checkmark  & \checkmark & -- & 93.17 & 98.83 \\
        Ours &  \checkmark  & \checkmark & \checkmark & \textbf{94.00} & \textbf{99.17} \\
        \bottomrule
    \end{tabular}
    \caption{Comprehensive evaluation of each stage in Cycle Training Framework.}
    \label{tab:stage_comparison}
\end{table}

\textbf{Table}~\ref{tab:stage_comparison} shows the results obtained from the three stages of our Cycle Training approach. In the initial stage (Exploitation), we directly leverage the pretrained backbone, which yields a Top-1 accuracy of 89.83\%. In the subsequent stage (Exploration), further fine-tuning of the model parameters boosts the Top-1 accuracy to 93.17\%, demonstrating the benefit of adapting the model to the target data. Finally, during the Stabilization stage, the model consolidates both the pretrained knowledge and the fine-tuned parameters to achieve the highest performance, with a Top-1 accuracy of 94.00\% and a Top-3 accuracy of 99.17\%. This progressive improvement across the stages clearly illustrates the effectiveness of our method.

\begin{table}[ht]
\centering
\footnotesize
\setlength{\tabcolsep}{4pt}
\begin{tabular}{lcccc}
\toprule
\textbf{Method} & \textbf{SSDA} & \textbf{Aug.} & \textbf{Top-1 (\%)} & \textbf{Top-3 (\%)}\\
\midrule
MobileNetv2             & --           & --           & 91.67  & 98.83\\
MobileNetv2 + Aug        & --           & \checkmark  & 92.00 & 98.83\\
MobileNetv2 + SSDA     & \checkmark   & --           & 92.83 & 98.83\\
MobileNetv2 + SSDA + Aug & \checkmark  & \checkmark  & \textbf{94.00} & \textbf{99.17S}\\
\bottomrule
\end{tabular}
\caption{Evaluation of Semi-Supervised Domain Adaptation and Augmentations.}
\label{tab:ssda_aug}
\end{table}

In \textbf{Table \ref{tab:ssda_aug}}, we observed that while the Top-3 Accuracy remains relatively unchanged across different configurations, the Top-1 Accuracy improves significantly with the addition of Augmentation or Pseudo Labeling. This suggests that the model is better at distinguishing between closely competing classes, particularly for samples that were previously ambiguous. In other words, although the correct class is still present among the top three predictions, the model now more confidently selects it as the top prediction, indicating enhanced discriminative capability. Furthermore, the best performance is achieved when both Augmentation and Pseudo Labeling are applied in combination, demonstrating that their synergy provides a more robust and accurate model.

\subsection{Local Execution on an Edge Device}

\begin{figure}[ht]
    \centering
    \begin{subfigure}{0.22\textwidth}
        \centering
        \includegraphics[width=\textwidth]{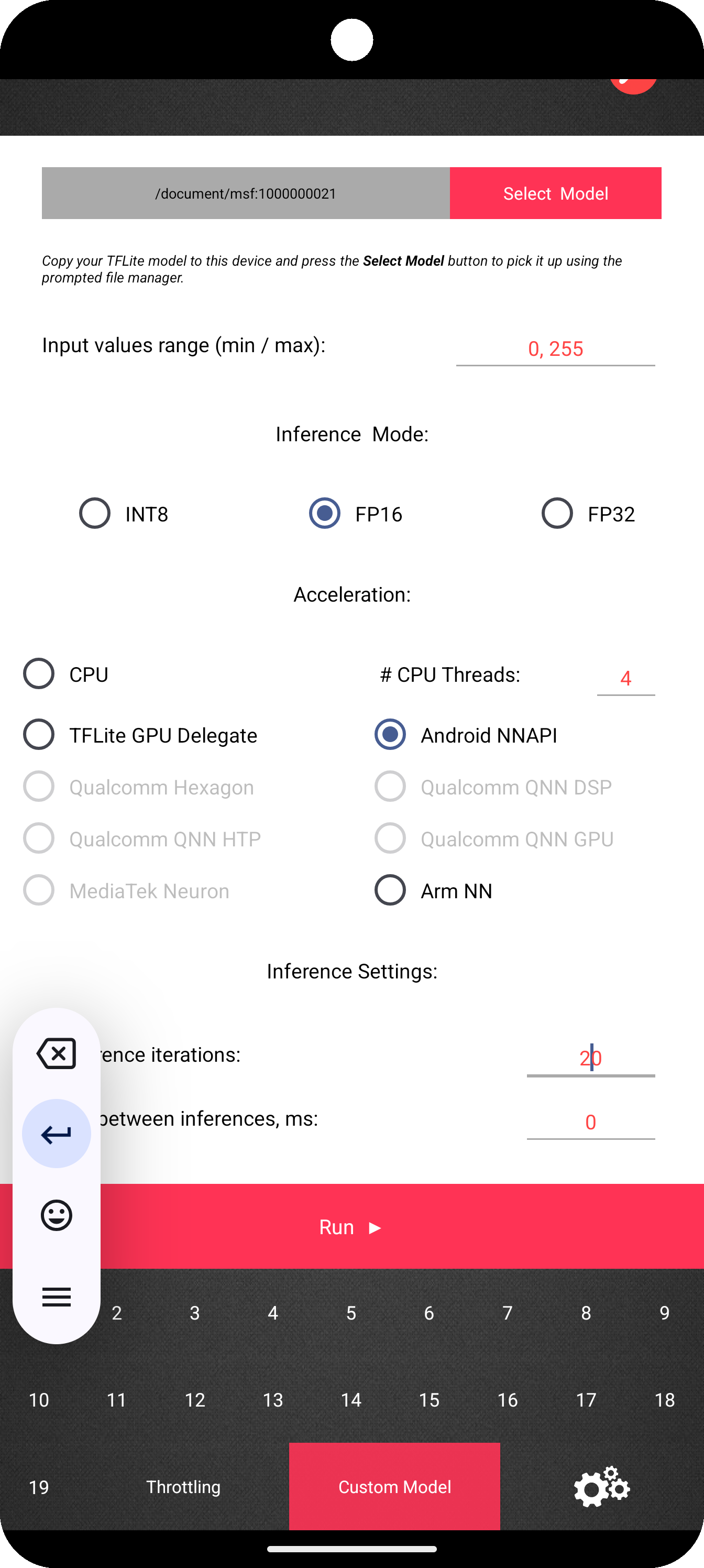}
        \label{fig:mobile1}
    \end{subfigure}
    \hfill
    \begin{subfigure}{0.22\textwidth}
        \centering
        \includegraphics[width=\textwidth]{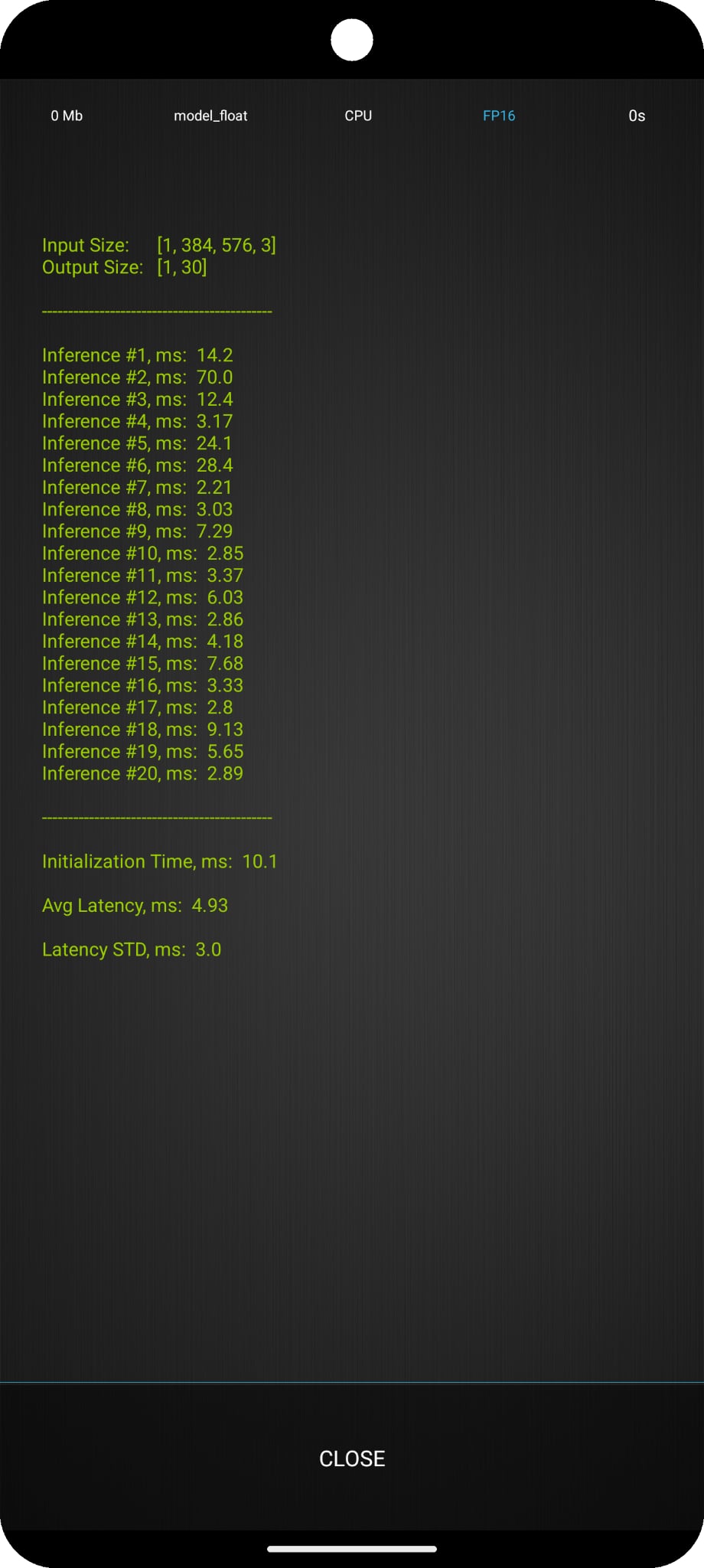}
        \label{fig:mobile2}
    \end{subfigure}
    \caption{Overall caption describing both subfigures.}
    \label{fig:mobile}
\end{figure}

In this competition, participants are tasked not only with developing models and techniques for image classification but also with executing these models on mobile or edge devices. To facilitate testing on mobile devices, the competition provides the AI Benchmark application. This tool allows users to load converted TensorFlow Lite models onto any device that supports GPU acceleration such as  TFLite GPU Delegate, Qualcomm Hexagon, Qualcomm QNN DSP. Qualcomm QNN HTP, Qualcomm QNN GPU, MediaTek Neuron, Android NNAPI, Arm NN. After completing the training process on an RTX A6000 GPU with 48GB of memory, the model is converted to TensorFlow Lite (TFLite) format, enabling its deployment and use on mobile devices. In the application, the trained model is loaded from the Download folder, with the acceleration backend configured to Android NNAPI and inference mode FP16. The number of inference iterations is set to 20 to ensure consistent performance evaluation. As illustrated in \textbf{Figure~\ref{fig:mobile}}, the model is evaluated on a Pixel phone, achieving a runtime of 4.93 ms with NNAPI, demonstrating efficient performance on a mobile device.

\subsection{Discussion}
The experimental results validate the effectiveness of our framework. The teacher network (ResNet101x3) effectively captures robust feature representations and benefits from pseudo labeling, while the student network (MobileNetv2) attains significant accuracy improvements through Cycle Training with tailored augmentation. Despite its compact size (14.85 MB), our MobileNetv2 model achieves a \textbf{top-1 accuracy of 94.00\%} and maintains real-time inference capabilities on mobile devices.

Moreover, our ablation studies confirm that the integration of semi-supervised domain adaptation and augmentation enhances the model's discriminative power. The progressive improvements observed across the Cycle Training stages further underscore the efficacy of our iterative training process. Comprehensive details on hyperparameters, TFLite conversion scripts, and training procedures are provided in the supplementary material to ensure full reproducibility.

%% file: sec/5_conclusion.tex
\section{Conclusion}
\label{sec:conclusion}

In conclusion, we presented a novel framework that integrates Semi-Supervised Domain Adaptation (SSDA) with a Cycle Training (CT) strategy to address the challenges of deploying deep learning models on mobile devices. By leveraging a high-capacity ResNet-101 (BiT) teacher model for high-confidence pseudo labeling and transferring its knowledge to a lightweight MobileNet-V2 student model, our approach effectively bridges the domain gap and enhances model performance. Comprehensive experiments on the CamSSD dataset demonstrate that our method achieves a Top-1 accuracy of 94.00\% and a Top-3 accuracy of 99.17\%, while maintaining a real-time inference speed of 1.61 ms on a mobile device. These results highlight the potential of our framework in deploying deep learning models efficiently on resource-limited hardware. 

%% file: main.bbl
\begin{thebibliography}{48}
\providecommand{\natexlab}[1]{#1}
\providecommand{\url}[1]{\texttt{#1}}
\expandafter\ifx\csname urlstyle\endcsname\relax
  \providecommand{\doi}[1]{doi: #1}\else
  \providecommand{\doi}{doi: \begingroup \urlstyle{rm}\Url}\fi

\bibitem[aiv()]{aivietnamVit}
{A}{I} {VIETNAM} --- aivietnam.edu.vn.

\bibitem[Ab~Wahab et~al.(2021)Ab~Wahab, Nazir, Ren, Noor, Akbar, and Mohamed]{ab2021efficientnet}
Mohd~Nadhir Ab~Wahab, Amril Nazir, Anthony Tan~Zhen Ren, Mohd Halim~Mohd Noor, Muhammad~Firdaus Akbar, and Ahmad Sufril~Azlan Mohamed.
\newblock Efficientnet-lite and hybrid cnn-knn implementation for facial expression recognition on raspberry pi.
\newblock \emph{IEEE Access}, 9:\penalty0 134065--134080, 2021.

\bibitem[Abadade et~al.(2023)Abadade, Temouden, Bamoumen, Benamar, Chtouki, and Hafid]{tinyml2}
Youssef Abadade, Anas Temouden, Hatim Bamoumen, Nabil Benamar, Yousra Chtouki, and Abdelhakim~Senhaji Hafid.
\newblock A comprehensive survey on tinyml.
\newblock \emph{IEEE Access}, 11:\penalty0 96892--96922, 2023.

\bibitem[Alkhulaifi et~al.(2021)Alkhulaifi, Alsahli, and Ahmad]{alkhulaifi2021knowledge}
Abdolmaged Alkhulaifi, Fahad Alsahli, and Irfan Ahmad.
\newblock Knowledge distillation in deep learning and its applications.
\newblock \emph{PeerJ Computer Science}, 7:\penalty0 e474, 2021.

\bibitem[Chebotar and Waters(2016)]{chebotar2016distilling}
Yevgen Chebotar and Austin Waters.
\newblock Distilling knowledge from ensembles of neural networks for speech recognition.
\newblock In \emph{Interspeech}, pages 3439--3443, 2016.

\bibitem[Cho and Hariharan(2019)]{know3}
Jang~Hyun Cho and Bharath Hariharan.
\newblock On the efficacy of knowledge distillation.
\newblock In \emph{Proceedings of the IEEE/CVF international conference on computer vision}, pages 4794--4802, 2019.

\bibitem[Dong et~al.(2020)Dong, Zhou, Ruan, and Li]{dong2020mobilenetv2}
Ke Dong, Chengjie Zhou, Yihan Ruan, and Yuzhi Li.
\newblock Mobilenetv2 model for image classification.
\newblock In \emph{2020 2nd International Conference on Information Technology and Computer Application (ITCA)}, pages 476--480. IEEE, 2020.

\bibitem[Gou et~al.(2021)Gou, Yu, Maybank, and Tao]{gou2021knowledge}
Jianping Gou, Baosheng Yu, Stephen~J Maybank, and Dacheng Tao.
\newblock Knowledge distillation: A survey.
\newblock \emph{International Journal of Computer Vision}, 129\penalty0 (6):\penalty0 1789--1819, 2021.

\bibitem[Han et~al.(2024)Han, Yuan, Wei, and Yu]{han2024regmixmatch}
Haorong Han, Jidong Yuan, Chixuan Wei, and Zhongyang Yu.
\newblock Regmixmatch: Optimizing mixup utilization in semi-supervised learning.
\newblock \emph{arXiv preprint arXiv:2412.10741}, 2024.

\bibitem[Han et~al.(2020)Han, Wang, Tian, Guo, Xu, and Xu]{han2020ghostnet}
Kai Han, Yunhe Wang, Qi Tian, Jianyuan Guo, Chunjing Xu, and Chang Xu.
\newblock Ghostnet: More features from cheap operations.
\newblock In \emph{Proceedings of the IEEE/CVF conference on computer vision and pattern recognition}, pages 1580--1589, 2020.

\bibitem[Han et~al.(2015{\natexlab{a}})Han, Mao, and Dally]{han2015deep}
Song Han, Huizi Mao, and William~J Dally.
\newblock Deep compression: Compressing deep neural networks with pruning, trained quantization and huffman coding.
\newblock \emph{arXiv preprint arXiv:1510.00149}, 2015{\natexlab{a}}.

\bibitem[Han et~al.(2015{\natexlab{b}})Han, Pool, Tran, and Dally]{han2015learning}
Song Han, Jeff Pool, John Tran, and William Dally.
\newblock Learning both weights and connections for efficient neural network.
\newblock \emph{Advances in neural information processing systems}, 28, 2015{\natexlab{b}}.

\bibitem[Han et~al.(2016)Han, Liu, Mao, Pu, Pedram, Horowitz, Dally, et~al.]{han2016deep}
Song Han, Xingyu Liu, Huizi Mao, Jing Pu, Ardavan Pedram, Mark Horowitz, Bill Dally, et~al.
\newblock Deep compression and eie: Efficient inference engine on compressed deep neural network.
\newblock In \emph{Hot Chips Symposium}, pages 1--6, 2016.

\bibitem[He et~al.(2016)He, Zhang, Ren, and Sun]{resnet}
Kaiming He, Xiangyu Zhang, Shaoqing Ren, and Jian Sun.
\newblock Deep residual learning for image recognition.
\newblock In \emph{Proceedings of the IEEE conference on computer vision and pattern recognition}, pages 770--778, 2016.

\bibitem[He et~al.(2017)He, Zhang, and Sun]{he2017channel}
Yihui He, Xiangyu Zhang, and Jian Sun.
\newblock Channel pruning for accelerating very deep neural networks.
\newblock In \emph{Proceedings of the IEEE international conference on computer vision}, pages 1389--1397, 2017.

\bibitem[Hinton et~al.(2015)Hinton, Vinyals, and Dean]{hinton2015distilling}
Geoffrey Hinton, Oriol Vinyals, and Jeff Dean.
\newblock Distilling the knowledge in a neural network.
\newblock \emph{arXiv preprint arXiv:1503.02531}, 2015.

\bibitem[Hluchyj and Karol(1991)]{hluchyj1991shuffle}
Michael~G Hluchyj and Mark~J Karol.
\newblock Shuffle net: An application of generalized perfect shuffles to multihop lightwave networks.
\newblock \emph{Journal of Lightwave Technology}, 9\penalty0 (10):\penalty0 1386--1397, 1991.

\bibitem[Howard et~al.(2017)Howard, Zhu, Chen, Kalenichenko, Wang, Weyand, Andreetto, and Adam]{howard2017mobilenets}
Andrew~G Howard, Menglong Zhu, Bo Chen, Dmitry Kalenichenko, Weijun Wang, Tobias Weyand, Marco Andreetto, and Hartwig Adam.
\newblock Mobilenets: Efficient convolutional neural networks for mobile vision applications.
\newblock \emph{arXiv preprint arXiv:1704.04861}, 2017.

\bibitem[Huang et~al.(2017)Huang, Liu, Van Der~Maaten, and Weinberger]{densetnet}
Gao Huang, Zhuang Liu, Laurens Van Der~Maaten, and Kilian~Q Weinberger.
\newblock Densely connected convolutional networks.
\newblock In \emph{Proceedings of the IEEE conference on computer vision and pattern recognition}, pages 4700--4708, 2017.

\bibitem[Huang et~al.(2023)Huang, Shen, Yu, Han, and Liu]{huang2023flatmatch}
Zhuo Huang, Li Shen, Jun Yu, Bo Han, and Tongliang Liu.
\newblock Flatmatch: Bridging labeled data and unlabeled data with cross-sharpness for semi-supervised learning.
\newblock \emph{Advances in Neural Information Processing Systems}, 36:\penalty0 18474--18494, 2023.

\bibitem[Ignatov et~al.(2021)Ignatov, Malivenko, and Timofte]{mobile2}
Andrey Ignatov, Grigory Malivenko, and Radu Timofte.
\newblock Fast and accurate quantized camera scene detection on smartphones, mobile ai 2021 challenge: Report.
\newblock In \emph{Proceedings of the IEEE/CVF Conference on computer vision and pattern recognition}, pages 2558--2568, 2021.

\bibitem[Kolesnikov et~al.(2020)Kolesnikov, Beyer, Zhai, Puigcerver, Yung, Gelly, and Houlsby]{kolesnikov2020big}
Alexander Kolesnikov, Lucas Beyer, Xiaohua Zhai, Joan Puigcerver, Jessica Yung, Sylvain Gelly, and Neil Houlsby.
\newblock Big transfer (bit): General visual representation learning.
\newblock In \emph{Computer Vision--ECCV 2020: 16th European Conference, Glasgow, UK, August 23--28, 2020, Proceedings, Part V 16}, pages 491--507. Springer, 2020.

\bibitem[Koonce(2021)]{koonce2021efficientnet}
Brett Koonce.
\newblock Efficientnet.
\newblock In \emph{Convolutional neural networks with swift for Tensorflow: image recognition and dataset categorization}, pages 109--123. Springer, 2021.

\bibitem[Koonce and Koonce(2021)]{koonce2021mobilenetv3}
Brett Koonce and Brett Koonce.
\newblock Mobilenetv3.
\newblock \emph{Convolutional Neural Networks with Swift for Tensorflow: Image Recognition and Dataset Categorization}, pages 125--144, 2021.

\bibitem[Krishnamoorthi(1806)]{krishnamoorthi1806quantizing}
Raghuraman Krishnamoorthi.
\newblock Quantizing deep convolutional networks for efficient inference: A whitepaper. arxiv 2018.
\newblock \emph{arXiv preprint arXiv:1806.08342}, 1806.

\bibitem[Krizhevsky et~al.(2012)Krizhevsky, Sutskever, and Hinton]{alexnet}
Alex Krizhevsky, Ilya Sutskever, and Geoffrey~E Hinton.
\newblock Imagenet classification with deep convolutional neural networks.
\newblock \emph{Advances in neural information processing systems}, 25, 2012.

\bibitem[Le-Quynh et~al.(2023)Le-Quynh, Nguyen, Quang-Hoang, Dinh, Nguyen, Ngo, and An]{10.1145/3628797.3629011}
Minh-Dung Le-Quynh, Anh-Tuan Nguyen, Anh-Tuan Quang-Hoang, Van-Huy Dinh, Tien-Huy Nguyen, Hoang-Bach Ngo, and Minh-Hung An.
\newblock Enhancing video retrieval with robust clip-based multimodal system.
\newblock In \emph{Proceedings of the 12th International Symposium on Information and Communication Technology}, page 972–979, New York, NY, USA, 2023. Association for Computing Machinery.

\bibitem[Lin et~al.(2023)Lin, Zhu, Chen, Wang, and Han]{tinyml3}
Ji Lin, Ligeng Zhu, Wei-Ming Chen, Wei-Chen Wang, and Song Han.
\newblock Tiny machine learning: Progress and futures [feature].
\newblock \emph{IEEE Circuits and Systems Magazine}, 23\penalty0 (3):\penalty0 8–34, 2023.

\bibitem[Moslemi et~al.(2024)Moslemi, Briskina, Dang, and Li]{moslemi2024survey}
Amir Moslemi, Anna Briskina, Zubeka Dang, and Jason Li.
\newblock A survey on knowledge distillation: Recent advancements.
\newblock \emph{Machine Learning with Applications}, page 100605, 2024.

\bibitem[Ngo et~al.(2024)Ngo, Lam, Nguyen, Dinh, and Choi]{ngo2024dual}
Ba~Hung Ngo, Ba~Thinh Lam, Thanh~Huy Nguyen, Quang~Vinh Dinh, and Tae~Jong Choi.
\newblock Dual dynamic consistency regularization for semi-supervised domain adaptation.
\newblock \emph{IEEE Access}, 2024.

\bibitem[Nguyen et~al.(2025)Nguyen, Vu, Duong, Duong, Nguyen, and Dinh]{nguyen2025enhancing}
Khoi~Anh Nguyen, Linh~Yen Vu, Thang~Dinh Duong, Thuan~Nguyen Duong, Huy~Thanh Nguyen, and Vinh~Quang Dinh.
\newblock Enhancing vietnamese vqa through curriculum learning on raw and augmented text representations.
\newblock \emph{arXiv preprint arXiv:2503.03285}, 2025.

\bibitem[Nguyen et~al.()Nguyen, Nguyen, Nguyen, Vu, Dinh, and MERIAUDEAU]{nguyensemi}
Thanh-Huy Nguyen, Thien Nguyen, Xuan~Bach Nguyen, Nguyen Lan~Vi Vu, Vinh~Quang Dinh, and Fabrice MERIAUDEAU.
\newblock Semi-supervised skin lesion segmentation under dual mask ensemble with feature discrepancy co-training.
\newblock In \emph{Medical Imaging with Deep Learning}.

\bibitem[Nguyen et~al.(2024{\natexlab{a}})Nguyen, Tran, and Quang-Hoang]{nguyen2024improvinggeneralizationvisualreasoning}
Tien-Huy Nguyen, Quang-Khai Tran, and Anh-Tuan Quang-Hoang.
\newblock Improving generalization in visual reasoning via self-ensemble, 2024{\natexlab{a}}.

\bibitem[Nguyen et~al.(2024{\natexlab{b}})Nguyen, Tran, Tran, Phan-Nguyen, and Nguyen]{10661057}
Tho-Quang Nguyen, Huu-Loc Tran, Tuan-Khoa Tran, Huu-Phong Phan-Nguyen, and Tien-Huy Nguyen.
\newblock Fa-yolov9: Improved yolov9 based on feature attention block.
\newblock In \emph{2024 International Conference on Multimedia Analysis and Pattern Recognition (MAPR)}, pages 1--6, 2024{\natexlab{b}}.

\bibitem[Nguyen et~al.(2024{\natexlab{c}})Nguyen, Nguyen, Nguyen, Do, and Dinh]{nguyen2024emotic}
Xuan-Bach Nguyen, Hoang-Thien Nguyen, Thanh-Huy Nguyen, Nhu-Tai Do, and Quang~Vinh Dinh.
\newblock Emotic masked autoencoder on dual-views with attention fusion for facial expression recognition.
\newblock In \emph{Proceedings of the IEEE/CVF Conference on Computer Vision and Pattern Recognition}, pages 4784--4792, 2024{\natexlab{c}}.

\bibitem[Park et~al.(2019)Park, Kim, Lu, and Cho]{know2}
Wonpyo Park, Dongju Kim, Yan Lu, and Minsu Cho.
\newblock Relational knowledge distillation.
\newblock In \emph{Proceedings of the IEEE/CVF conference on computer vision and pattern recognition}, pages 3967--3976, 2019.

\bibitem[Pl{\"o}tz and Guan(2018)]{plotz2018deep}
Thomas Pl{\"o}tz and Yu Guan.
\newblock Deep learning for human activity recognition in mobile computing.
\newblock \emph{Computer}, 51\penalty0 (5):\penalty0 50--59, 2018.

\bibitem[Pouget et~al.(2021)Pouget, Ramesh, Giang, Chandrapalan, Tanner, Prussing, Timofte, and Ignatov]{pouget2021fast}
Angeline Pouget, Sidharth Ramesh, Maximilian Giang, Ramithan Chandrapalan, Toni Tanner, Moritz Prussing, Radu Timofte, and Andrey Ignatov.
\newblock Fast and accurate camera scene detection on smartphones.
\newblock In \emph{Proceedings of the IEEE/CVF Conference on Computer Vision and Pattern Recognition}, pages 2569--2580, 2021.

\bibitem[Qin et~al.(2024)Qin, Leichner, Delakis, Fornoni, Luo, Yang, Wang, Banbury, Ye, Akin, et~al.]{qin2024mobilenetv4}
Danfeng Qin, Chas Leichner, Manolis Delakis, Marco Fornoni, Shixin Luo, Fan Yang, Weijun Wang, Colby Banbury, Chengxi Ye, Berkin Akin, et~al.
\newblock Mobilenetv4: universal models for the mobile ecosystem.
\newblock In \emph{European Conference on Computer Vision}, pages 78--96. Springer, 2024.

\bibitem[Ray(2022)]{tinyml1}
Partha~Pratim Ray.
\newblock A review on tinyml: State-of-the-art and prospects.
\newblock \emph{Journal of King Saud University-Computer and Information Sciences}, 34\penalty0 (4):\penalty0 1595--1623, 2022.

\bibitem[Sandler et~al.(2018)Sandler, Howard, Zhu, Zhmoginov, and Chen]{sandler2018mobilenetv2}
Mark Sandler, Andrew Howard, Menglong Zhu, Andrey Zhmoginov, and Liang-Chieh Chen.
\newblock Mobilenetv2: Inverted residuals and linear bottlenecks.
\newblock In \emph{Proceedings of the IEEE conference on computer vision and pattern recognition}, pages 4510--4520, 2018.

\bibitem[Simonyan and Zisserman(2014)]{vggnet}
Karen Simonyan and Andrew Zisserman.
\newblock Very deep convolutional networks for large-scale image recognition.
\newblock \emph{arXiv preprint arXiv:1409.1556}, 2014.

\bibitem[Sohn et~al.(2020)Sohn, Berthelot, Carlini, Zhang, Zhang, Raffel, Cubuk, Kurakin, and Li]{sohn2020fixmatch}
Kihyuk Sohn, David Berthelot, Nicholas Carlini, Zizhao Zhang, Han Zhang, Colin~A Raffel, Ekin~Dogus Cubuk, Alexey Kurakin, and Chun-Liang Li.
\newblock Fixmatch: Simplifying semi-supervised learning with consistency and confidence.
\newblock \emph{Advances in neural information processing systems}, 33:\penalty0 596--608, 2020.

\bibitem[Szegedy et~al.(2015)Szegedy, Liu, Jia, Sermanet, Reed, Anguelov, Erhan, Vanhoucke, and Rabinovich]{inception}
Christian Szegedy, Wei Liu, Yangqing Jia, Pierre Sermanet, Scott Reed, Dragomir Anguelov, Dumitru Erhan, Vincent Vanhoucke, and Andrew Rabinovich.
\newblock Going deeper with convolutions.
\newblock In \emph{Proceedings of the IEEE conference on computer vision and pattern recognition}, pages 1--9, 2015.

\bibitem[Wang et~al.(2022)Wang, Chen, Heng, Hou, Fan, Wu, Wang, Savvides, Shinozaki, Raj, et~al.]{wang2022freematch}
Yidong Wang, Hao Chen, Qiang Heng, Wenxin Hou, Yue Fan, Zhen Wu, Jindong Wang, Marios Savvides, Takahiro Shinozaki, Bhiksha Raj, et~al.
\newblock Freematch: Self-adaptive thresholding for semi-supervised learning.
\newblock \emph{arXiv preprint arXiv:2205.07246}, 2022.

\bibitem[Yang et~al.(2019)Yang, Zhang, Zhang, Jin, and Chen]{yang2019essence}
Zhenchuan Yang, Chun Zhang, Weibin Zhang, Jianxiu Jin, and Dongpeng Chen.
\newblock Essence knowledge distillation for speech recognition.
\newblock \emph{arXiv preprint arXiv:1906.10834}, 2019.

\bibitem[Zhang et~al.(2021)Zhang, Wang, Hou, Wu, Wang, Okumura, and Shinozaki]{zhang2021flexmatch}
Bowen Zhang, Yidong Wang, Wenxin Hou, Hao Wu, Jindong Wang, Manabu Okumura, and Takahiro Shinozaki.
\newblock Flexmatch: Boosting semi-supervised learning with curriculum pseudo labeling.
\newblock \emph{Advances in neural information processing systems}, 34:\penalty0 18408--18419, 2021.

\bibitem[Zhang et~al.(2017)Zhang, Cisse, Dauphin, and Lopez-Paz]{zhang2017mixup}
Hongyi Zhang, Moustapha Cisse, Yann~N Dauphin, and David Lopez-Paz.
\newblock mixup: Beyond empirical risk minimization.
\newblock \emph{arXiv preprint arXiv:1710.09412}, 2017.

\end{thebibliography}
